\documentclass{article}
\usepackage{spconf,amsmath,graphicx,url}

\title{CSfM: Community-based Structure from Motion}
\name{Hainan Cui$^1$\sthanks{Corresponding author: hncui@nlpr.ia.ac.cn, shshen@nlpr.ia.ac.cn}, Shuhan Shen$^{1,2}$, Xiang Gao$^{1,2}$, Zhanyi Hu$^{1,2}$
\thanks{This work was supported by the Natural Science Foundation of China under Grants 61333015, 61421004 and 61473292.}}
\address{1. NLPR, Institute of Automation, Chinese Academy of Sciences, Beijing 100190, P. R. China\\
  2. University of Chinese Academy of Sciences, Beijing 100149, P. R. China }

\begin{document}
%\ninept
%
\maketitle
\begin{abstract}
Structure-from-Motion approaches could be broadly divided into two classes: incremental and global.
While incremental manner is robust to outliers, it suffers from error accumulation and heavy computation load.
The global manner has the advantage of simultaneously estimating all camera poses, but it is usually sensitive to epipolar geometry outliers.
In this paper, we propose an adaptive community-based SfM (CSfM) method which takes both robustness and efficiency into consideration.
First, the epipolar geometry graph is partitioned into separate communities.
Then, the reconstruction problem is solved for each community in parallel.
Finally, the reconstruction results are merged by a novel global similarity averaging method, which solves three convex $L1$ optimization problems.
Experimental results show that our method performs better than many of the state-of-the-art global SfM approaches
in terms of computational efficiency, while achieves similar or better reconstruction accuracy and robustness than
many of the state-of-the-art incremental SfM approaches.

\end{abstract}
\newcommand{\etal}{\emph{et al.}}
\newcommand{\eg}{\emph{e.g.\ }}
\newcommand{\ie}{\emph{i.e.\ }}
\section{Introduction}\label{sec:intro}
Structure-from-Motion (SfM) technique is used to simultaneously estimate the 3D scene points and camera poses from a collection of images.
Based on the manner of initial camera poses estimation, SfM approaches could be divided into two classes: incremental and global.

Incremental SfM methods~\cite{Snavely_Bundler, VSFM, colmap} compute the camera poses sequentially, which are robust
at computing an accurate 3D model due to many trials to remove outliers via RANSAC and bundle adjustment~\cite{triggs2000bundle}.
However for large-scale reconstruction applications, such incremental manner suffers from heavy computational
load because the time-consuming bundle adjustment is repeatedly performed,
and error accumulation may cause scene drift~\cite{Drift}.
In comparison, global SfM~\cite{1DSfM, openMVG_L, SfM2, SfM1, cui2015global} methods
simultaneously compute camera poses from the epipolar geometry graph (EG),
where vertices correspond to images and edges link matched image pairs, and only perform the bundle adjustment once.
However, those essential matrix based methods are more sensitive to the epipolar geometry outliers,
and they could only calibrate images in the parallel rigid graph~\cite{ozyesil2015robust}.
As a result, many useful images are likely to be discarded because the EG
may not be dense and accurate enough.
Considering both efficiency and robustness, some work~\cite{Snavely_Skeletal, shah2014multistage} used
only a subset of images in the incremental reconstruction. For example, Snavely \etal~\cite{Snavely_Skeletal} constructed a skeletal
graph first, then incrementally reconstructed the images in this skeletal graph to get a coarse
scene reconstruction, finally based on the 2D-3D matches, the other images are registered.
However, the graph construction process in these methods is sometimes time-consuming, and the problem of
finding iconic images is non-trivial.

The traditional SfM methods usually consider all the images as a sole community,
but for large-scale scene reconstructions,
especially for those unordered images searched from Internet~\cite{1DSfM}, the images distribution usually has a community-character:
some interested places get denser images, while sparser for other places.
In such case, simultaneously reconstructing all the images is not a rational choice.
In comparison, hierarchical SfM methods~\cite{havlena2009randomized, Havlena2010, toldo2015hierarchical} are to
create atomic models first, and then incrementally merge different models.
However, such a manner is sensitive to the atomic model chosen and model growing manner.
Bhowmick \etal~\cite{Divide_Conquer} used the vocabulary tree to cluster the scene by the normalized cuts.
While efficient, it can neither determine whether the image data has a community-character, nor
automatically determine the number of communities. Besides, the scene completeness could not be guaranteed by clustering on a
coarse EG produced by the image-retrieval.
Thus, in this paper we propose an automatic manner to detect whether the image data has a community-character, and if yes,
the images are divided into independent communities first, and then the reconstruction is performed for each community in parallel,
followed by a merging step to align all the reconstruction results into a united global frame.

\begin{figure}
\centerline{\includegraphics[width=0.4\textwidth]{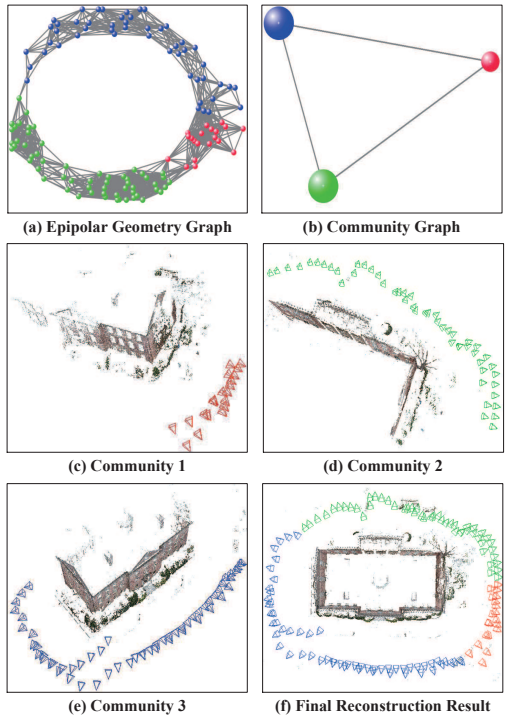}}
\caption{Reconstruction result on dataset Building~\cite{zach2010disambiguating}.}
\label{introResult}
\end{figure}
Our \textbf{Contributions} include:
 (1) an automatic community detection method is proposed to determine whether the image dataset should be divided into communities;
 (2) a community graph constructing method is proposed to divide the epipolar geometry graph into groups with denser connections inside
 and sparser connections outside;
 (3) a novel global similarity averaging method is proposed to merge all the separate reconstruction results
 in each community into a united global frame,
 where three convex $L1$ optimization problems are solved, and followed by a bundle adjustment step to
 further refine all the camera poses and reconstructed scene.

Fig.~\ref{introResult} illustrates the reconstruction result using our proposed method on a public dataset Building~\cite{zach2010disambiguating}.
We can see that the epipolar geometry graph is automatically divided into 3 communities, and each community captures a part of the scene.
In this way, each community could be reconstructed in parallel, and the time-cost of our reconstruction is the maximal running-time
spent among all these communities.
After a global similarity averaging, these separate reconstruction results are merged into a final reconstruction result,
which is showed in Fig.~\ref{introResult}(f).
With a comparable calibration accuracy, the speed of our CSfM is 10 times faster than Bundler~\cite{Snavely_Bundler},
and 4 times faster than COLMAP~\cite{colmap}.
Extensive experiments in Sec.~\ref{exp} show that our method has a larger scalability than many of the state-of-the-art SfM methods.

\section{Community Detection}
Given the epipolar geometry graph (EG), whose vertices correspond to images and edges link matched image pairs,
our goal is to detect whether the image data has a community-character and how many communities exist.
Community detection has been widely
used in the complex networks analysis~\cite{Fortunato2010Community, Clauset2005Finding}, which aims to divide
a graph into subgraphs with denser connections inside and sparser connections outside.

Let $A_{ij}$ be an element of the adjacent matrix of EG. If camera $i$ and camera $j$ are connected, then $A_{ij} = 1$, otherwise $A_{ij} = 0$.
Let $d_i = \Sigma_jA_{ij}$ be the degree of vertex $i$ in the EG, which denotes the number of cameras connected to the $i^{th}$ camera, and $m = \frac{1}{2}\sum_{ij}A_{ij}$ be the number of edges in the EG.
Then, if the existence of the epipolar edge is randomized, the existence probability of an edge connecting camera $i$ and camera $j$ is
$\frac{d_id_j}{2m}$. To measure the difference of the fraction of intra-community connections between random graph and the
EG, we use the modularity indicator $Q$ proposed in ~\cite{Clauset2005Finding}.
Suppose that camera $i$ belongs to the community $U_p$ and camera $j$ belongs to the community $U_q$, then $Q$ is defined as:
\begin{equation}\label{E1}
\begin{array}{cl}
Q = \frac{1}{2m}\sum_{ij}(A_{ij} - \frac{d_id_j}{2m})\delta(U_p, U_q),
\end{array}
\end{equation}
where $\delta(U_p, U_q) = 1$ if $U_p$ = $U_q$ and 0 otherwise.
To divide the EG, each node is assumed belongs to a sole community first, then separate communities are iteratively joined when the amalgamations result in the largest increase in $Q$.
As proposed in~\cite{Clauset2005Finding}, the modularity has a single peak $Q_{max}$ over the generation of the dendrogram which indicates the most significant community structure. In practice, we found that $Q_{max} > 0.3$ indicates that EG has a significant community structure.
Thus, we take the partition result when the peaking value of $Q$ is larger than 0.3. If $Q_{max} < 0.3$, all the images should be considered as a single community.

However, for large-scale scene reconstruction problems, one-time partition is usually not enough because
some of the current partitions still possess a community-character.
Thus, we propose to iteratively divide the EG until each of the partition results could be considered as a single community.
When the iteration is finished, we get a community graph whose vertices correspond to communities and edges link communities.
Note that a pair of communities are linked when there are some epipolar edges between them.
For example, Fig.~\ref{introResult}(b) shows a community graph composed of three connected communities.
Furthermore, considering the robustness of scene reconstruction,
each community should reconstruct enough 3D points for the subsequent merging step,
thus we set a minimal number of images $N$ for each community (In our work, $N$ is set to 20).
After the iteration converges, those small datasets whose number of images is less than $N$
are merged into their closest connected communities.
The degree of the closeness between two communities is defined as the number of epipolar geometry edges across them.

\section{Global Similarity Averaging}
After community detection, the image connections become denser in each community, thus many SfM methods
could be used for reconstruction.
Given the co-visible 3D points between a pair of connected communities, RANSAC technique is used to
estimate a 3D similarity transformation~\cite{horn1987closed}, including relative scale-ratio,
rotation and translation.
However, communities usually construct a connected graph, thus we propose a global similarity averaging method to robustly align
those separate reconstruction results.

\subsection{Global Scale Averaging}
Let $s_{i}$ be the scale of the $i^{th}$ community, $s_{ij}$ be the scale factor in the 3D similarity
transformation between communities $U_i$ and $U_j$. Given $\frac {s_{i}} {s_{j}} = s_{ij}$, by taking log of both sides, we have
\begin{align}
log(s_{i}) - log(s_{j}) = log(s_{ij}).
\end{align}
By stacking the above equation from all the connected community pairs, we have a linear equation system:
$
\textbf{A}_s*\textbf{x}_s = \textbf{b}_s
$,
where $\textbf{x}_s$ and $\textbf{b}_s$ are the vectors by concatenating $log(s_{i})$ and $log(s_{ij})$ respectively,
and $\textbf{A}_s$ is a sparse matrix where nonzero values are only $1$ and $-1$.
As the scale estimation is up to global scale, to remove the gauge ambiguity,
we set the scale of first community $s_{1}$ as unit: $log(s_{1}) = 0$.
Then, the equation system is solved by the following L1 optimization~\cite{candes2005l1}:
$\arg \min \| \textbf{A}_s*\textbf{x}_s - \textbf{b}_s \|_{L1}$.
Note that this $L1$ optimization is convex and achieves the global optimum
as studied in~\cite{parikh2014proximal}.

\subsection{Global Rotation Averaging}
Let $\textbf{R}_{i}$ be the rotation transformation of the $i^{th}$ community,
$\textbf{R}_{ij}$ be the relative rotation transformation in the 3D similarity
transformation between communities $U_i$ and $U_j$.
Given $\textbf{R}_{ij} = \textbf{R}_j * \textbf{R}_i ^ T$, by taking log of both sides,
we get an equation between their corresponding angle-axis vectors:
\begin{align}
\textbf{w}_{ij} = \textbf{w}_j - \textbf{w}_i,
\end{align}
where $\textbf{w}_{ij}$ is the angle-axis vector corresponds to $\textbf{R}_{ij}$, and $\textbf{w}_{i}, \textbf{w}_{j}$ are the angle-axis vectors
correspond to $\textbf{R}_{i}$ and $\textbf{R}_{j}$ respectively.
By stacking the above equation from all connected community pairs,
we have a linear equation system. Then, the L1RA~\cite{chatterjee2013efficient} method is used for our rotation averaging.

Then for the $i^{th}$ community, given its scale $s_i$ and rotation transformation $\textbf{R}_i$,
we transform its $k^{th}$ scene point $\textbf{X}_{ik}$ by $s_i * \textbf{R}_i * \textbf{X}_{ik}$.
After this transformation, the translation $\textbf{T}_{ij}$ between a pair of connected communities is recomputed~\cite{horn1987closed}.

\subsection{Global Translation Averaging}
Let $\textbf{T}_{i}$ be the translation transformation of the $i^{th}$ community, $\textbf{T}_{ij}$ be the relative translation
transformation in the 3D similarity transformation between communities $U_i$ and $U_j$.
Similarly, by stacking the equation $\textbf{T}_{ij} = \textbf{T}_{j} - \textbf{T}_{i}$ from all
connected community pairs, we have a linear equation system.
Then, this equation system is solved by the following L1 optimization~\cite{candes2005l1}:
$\arg \min \| \textbf{A}_t*\textbf{x}_t - \textbf{b}_t \|_{L1}$,
where $\textbf{x}_t$ and $\textbf{b}_t$ are the vectors by concatenating $\textbf{T}_{i}$ and $\textbf{T}_{ij}$ respectively,
and $\textbf{A}_t$ is a sparse matrix where nonzero values are only $1$ and $-1$.
As the translation estimation is up to a global translation transformation, to remove the gauge ambiguity,
we set the translation of the first community as zero, $\textbf{T}_{1} = \textbf{0}$.
%\begin{figure*}
%\centerline{\includegraphics[width=0.8\textwidth]{Exp.eps}}
%\caption{Reconstruction result on dataset Building~\cite{zach2010disambiguating} and Campus~\cite{cui2015global}.}
%\label{fig:sequential}
%\end{figure*}
\subsection{Reconstructions Merging}
Given the scale, rotation, translation transformation for each community, the reconstruction results (3D scene points and camera poses)
of all the communities are merged into a single global frame.
For the $i^{th}$ community, its $k^{th}$ 3D scene point $\textbf{X}_{ik}^{o}$ is transformed by:
\begin{align}
\textbf{X}_{ik}^{g} = \textbf{X}_{ik}^{o} + \textbf{T}_i.
\end{align}

For the $j^{th}$ camera in the $i^{th}$ community, its camera rotation $\textbf{R}_{ij}^{o}$ and camera center $\textbf{C}_{ij}^{o}$ are transformed by:
\begin{align}
\textbf{R}_{ij}^{g} = \textbf{R}_{ij}^{o} * \textbf{R}_i^T, \\
\textbf{C}_{ij}^{g} = s_i * \textbf{R}_i * \textbf{C}_{ij}^{o} + \textbf{T}_i,
\end{align}
where $\textbf{X}_{ik}^{g}, \textbf{R}_{ij}^{g}, \textbf{C}_{ij}^{g}$ denote the transformed 3D scene point,
camera rotation, and camera center respectively in the final united global frame.
After merging, a final bundle adjustment is performed to further refine all the camera poses and scene points.
\begin{figure}
\centerline{\includegraphics[width=0.5\textwidth]{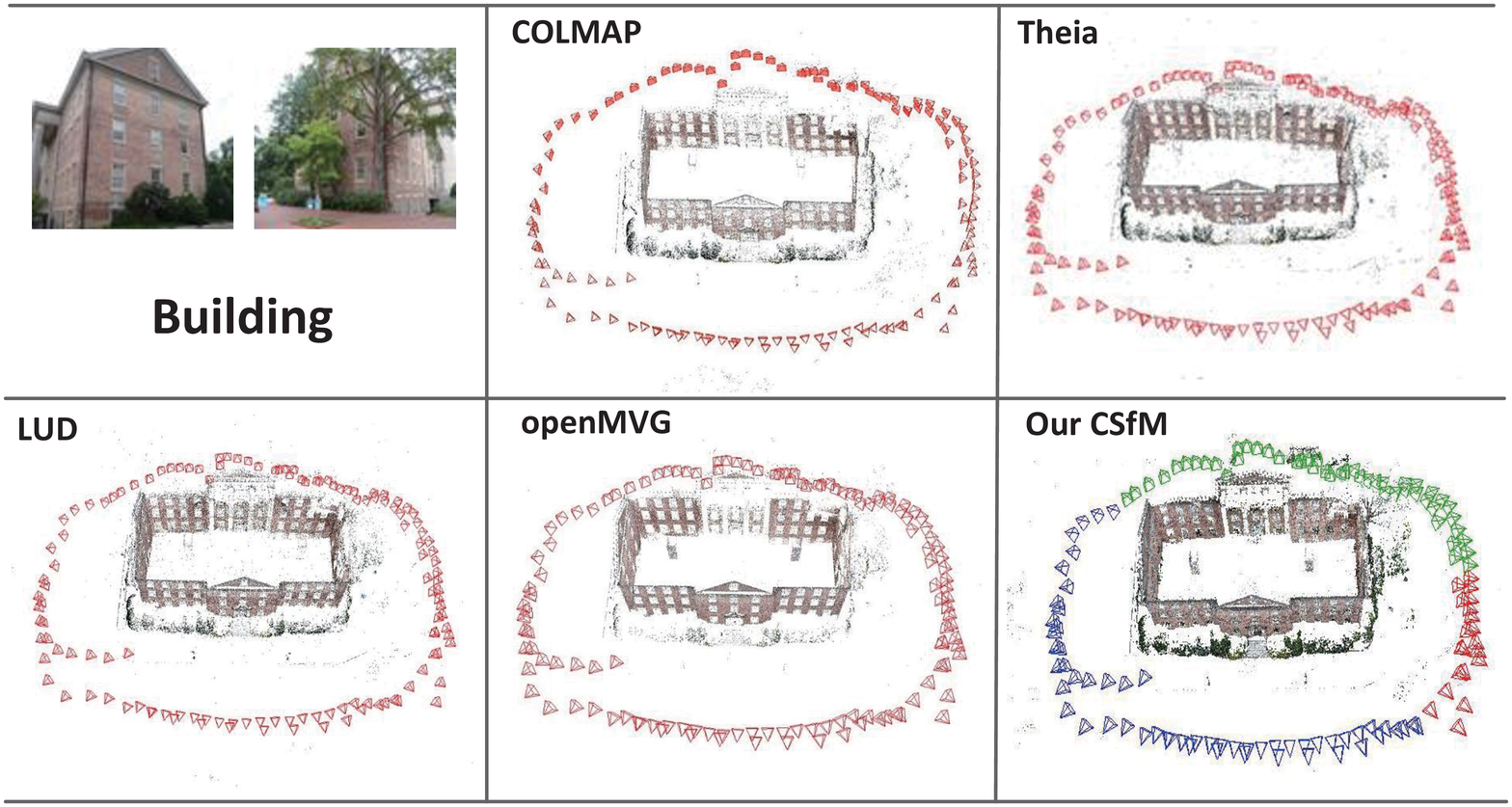}}
\caption{Reconstruction results on dataset Buidling~\cite{zach2010disambiguating}.}
\label{fig:seq1}
\end{figure}
\begin{figure}
\centerline{\includegraphics[width=0.5\textwidth]{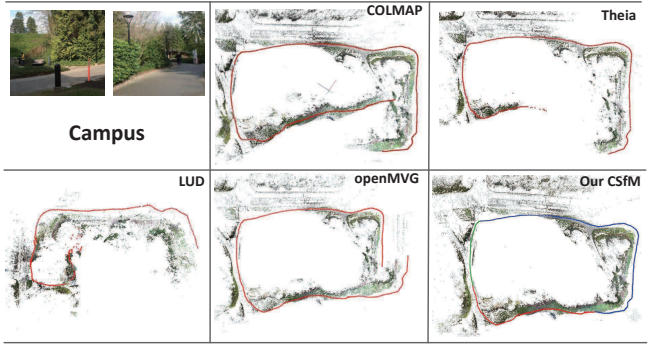}}
\caption{Reconstruction results on dataset Campus~\cite{cui2015global}.}
%\caption{Reconstruction result on dataset RomanForum~\cite{1DSfM} and ArtsQuad~\cite{Crandall_MRF2013}.}
\label{fig:seq2}
\end{figure}
\section{Experiments}{\label{exp}}
We validate our reconstruction method (CSfM) on both sequential and unordered image datasets.
All the experiments are performed on a PC with an Intel Xeon E5-2603 2.50GHz CPU (4 cores) and 32G RAM.
The optimization package Ceres-solver~\cite{ceres-solver} is used for our bundle adjustment.
Since the camera connectedness is dense in each community, many SfM methods could achieve similar reconstruction results for the same community.
Considering the robustness, we choose the state-of-the-art incremental SfM method COLMAP~\cite{colmap} to reconstruct in each community.

\subsection{Reconstruction on Sequential Image Data}
We evaluate our community-based SfM (CSfM) method on two public sequential image datasets: the \textbf{\textit{Building}}
with 128 images from~\cite{zach2010disambiguating}, whose modularity value $Q_{max}$ is 0.52;
the \textbf{\textit{Campus}} with 1040 images from~\cite{cui2015global}, whose modularity value $Q_{max}$ is 0.62.
We compare our method with two state-of-the-art incremental SfM methods: COLMAP~\cite{colmap} and Theia~\cite{Theia},
and two state-of-the-art global SfM methods: openMVG~\cite{openMVG_L} and LUD~\cite{ozyesil2015robust} .

For \textbf{\textit{Building}}, the reconstruction results comparison are shown in Fig.~\ref{fig:seq1}, from which we can see the
images are clustered into 3 communities using our proposed CSfM method, showing with different colored camera poses.
Since the EG is clean and images with enough connections,
all the methods in comparison could correctly reconstruct the scene. However, the time-cost of these methods are:
COLMAP 232s, Theia 363s, openMVG 381s, LUD 126s, our CSfM 59s, from which we can see our method is four times faster than COLMAP.

For \textbf{\textit{Campus}}, the reconstruction results comparison are shown in Fig.~\ref{fig:seq2}.
From the sample images, we can see that many trees exist in the scene, which makes the EG sparse
and contaminated by many feature matching outliers.
From the reconstruction results, we can see that conventional incremental methods~\cite{colmap, Theia} suffer from scene drift, and the
loop closure cannot be achieved. For global methods~\cite{openMVG_L, ozyesil2015robust}, the reconstruction results are also erroneous,
indicating that the global SfM methods are sensitive to the epipolar geometry outliers.
For our method, 3 communities are detected. By incremental reconstructing inside and global similarity averaging outside,
the error accumulation is largely decreased and the loop closure is achieved.
\begin{figure}
\centerline{\includegraphics[width=0.5\textwidth]{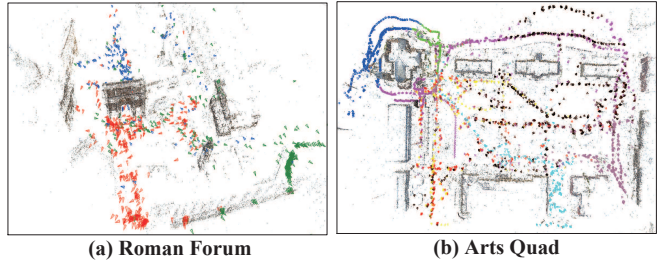}}
%\caption{Reconstruction result on dataset Campus~\cite{cui2015global}.}
\caption{Reconstruction results on dataset RomanForum, which is public in ~\cite{1DSfM}, and ArtsQuad which is public in~\cite{Crandall_MRF2013}.}
\label{fig:unordered}
\end{figure}

\begin{table}\small
\begin{center}
\caption{Accuracy and time-cost comparison on ArtsQuad.}
\label{table:Evaluation}
\begin{tabular}{|c|c|c|c|c|}
\hline
{} & {Bundler~\cite{Snavely_Bundler}} & {DISCO} & {COLMAP} & {Our method}\\
\hline
Accuracy & 1.01m & 1.16m & 0.85m & 0.83m \\
\hline
Time-cost & 62hrs & 7.7hrs & 3.5hrs & 0.6hrs \\
\hline
\end{tabular}
\end{center}
\end{table}

\subsection{Reconstruction on Unordered Image Data}
We evaluate our system on two large-scale unordered image datasets: \textit{\textbf{RomanForum}} and \textit{\textbf{ArtsQuad}}.
For \textit{\textbf{RomanForum}} with 1134 images, which is public in~\cite{1DSfM}, the corresponding modularity value $Q_{max}$ is 0.51.
It is divided into 3 communities, the corresponding camera poses with different
color are shown in the Fig.~\ref{fig:unordered}$(a)$,
from which we can see the cameras distribution is drastically uneven, and image connections become denser after community clustering.
For \textit{\textbf{ArtsQuad}} with 6514 images, which is public in DISCO~\cite{Crandall_MRF2013}, the modularity value $Q_{max}$ is 0.69.
This dataset is divided into 9 connected communities,
and its corresponding camera poses with different color are shown in the Fig.~\ref{fig:unordered}$(b)$.

For the comparison, we evaluate different reconstruction methods on \textit{\textbf{ArtsQuad}}~\cite{Crandall_MRF2013},
which has 348 ground-truth camera positions
measured by the differential GPS (with an accuracy about 10cm).
The result are shown in Table~\ref{table:Evaluation}, from which we can see that we achieve a similar median position accuracy
compared with other state-of-the-art methods, while the reconstruction efficiency is
greatly improved.
The results indicate that our method is more suitable for large-scale scene reconstructions since
the community-character is generically more common for the large-scale scenes.

In conclusion, our community-based SfM (CSfM) method inherits the robustness of incremental method and the efficiency of parallel reconstruction.

\section{Conclusion}{\label{conclusion}}
In this paper, we propose a community-based SfM method to tackle the reconstruction robustness and efficiency problems.
By clustering images into communities, the connections become denser inside and sparser outside.
The smaller-sized communities are reconstructed in parallel to substantially boost the computational efficiency and partial-scene reconstruction
quality.
Extensive reconstruction results show that our method performs better than many of the state-of-the-art SfM methods,
in terms of both efficiency and accuracy, and it is more suitable for large-scale scene reconstruction.

\bibliographystyle{IEEEbib}
\bibliography{CHN}

\end{document}